\documentclass{article}
\usepackage{ijcai19}

\usepackage{times}
\usepackage{soul}
\usepackage{url}
\usepackage[hidelinks]{hyperref}
\usepackage[utf8]{inputenc}
\usepackage[small]{caption}
\usepackage{graphicx}
\usepackage{amsmath}
\usepackage{booktabs}
\usepackage{algorithm}
\usepackage{algorithmic}
\usepackage{amssymb} 
\usepackage{color} 
\usepackage{mathrsfs}
\usepackage{makecell}
\usepackage{natbib}
\usepackage{stfloats} 

\title{Towards Understanding Adversarial Examples Systematically:\\Exploring Data Size, Task and Model Factors}

\author{
	Ke Sun$^1$
	\and
	Zhanxing Zhu$^{1,2}$ \footnote{Corresponding author.}\and
	Zhouchen Lin$^{3,4}$
	\affiliations
	$^1$Center for Data Science, Peking University, China\\
	$^2$Beijing Institute of Big Data Research (BIBDR), China\\
	$^3$Key Laboratory of Machine Perception (MOE), School of EECS, Peking University, China\\
	$^4$Cooperative Medianet Innovation Center, Shanghai Jiaotong University, China
	\emails
	\{ajksunke, zhanxing.zhu, zlin\}@pku.edu.cn
}
\begin{document}
	
	\maketitle
	
	\begin{abstract}
		Most previous works usually explained adversarial examples from several specific perspectives, lacking relatively integral comprehension about this problem. In this paper, we present a systematic study on adversarial examples from three aspects: the amount of training data, task-dependent  and model-specific factors. Particularly, we show that adversarial generalization~(i.e. test accuracy on adversarial examples) for \emph{standard training} requires more data than standard generalization~(i.e. test accuracy on clean examples); and uncover the global relationship between generalization and robustness with respect to the data size especially when data is augmented by generative models. This reveals the trade-off correlation between standard generalization and robustness in limited training data regime and their consistency when data size is large enough. Furthermore, we explore how different task-dependent and model-specific factors influence the vulnerability of deep neural networks by extensive empirical analysis. Relevant recommendations on defense against adversarial attacks are provided as well. Our results outline a potential path towards the luminous and systematic understanding of adversarial examples.
	\end{abstract}
	
	\section{Introduction}
	Although deep learning has achieved impressive performance in a wide range of machine learning tasks, recent research~\citep{szegedy2013intriguing,goodfellow6572explaining} has discovered that existing deep neural networks are susceptible to imperceptible perturbations of the input data, making erroneous but high-confident predictions. Furthermore, this phenomenon under the name of \textit{adversarial examples} is demonstrated ubiquitous in machine learning systems, causing great real-world security concerns.
	
	There has been a flurry of recent papers proposing adversarial attacks~\citep{moosavi2016deepfool,kurakin2016adversarial,carlini2017towards,madry2017towards,athalye2018obfuscated} and defenses~\citep{buckman2018thermometer,tramer2017ensemble,samangouei2018defense,madry2017towards,papernot2016distillation,liao2017defense} about this issue. Therefore, intelligent attacks against intelligent defenses become an arm race~\citep{wang2018one}. Apart from these, many hypotheses have been suggested in the literature, trying to explain the existence of adversarial examples from different perspectives. Linearity hypothesis~\citep{goodfellow6572explaining} was firstly proposed to explain this problem and obtained great acceptance. Later work~\citep{tanay2016boundary} studied the linearity hypothesis further and argued that adversarial examples exist when the classification boundaries lie close to the manifold of sampled data. \cite{su2018robustness} empirically found out the trade-off of accuracy and robustness and revealed that the robustness may be the cost of accuracy.
	
	All the aforementioned explanations are mostly proposed from specific perspectives to explain adversarial examples and there is hardly any work that can provide us a systematic understanding towards this phenomenon. In addition, \cite{schmidt2018adversarially} stated that adversarial robust generalization for \emph{adversarial training} requires more data and the data set may not be large enough for adversarial training to obtain a high robust generalization. A natural question could be raised: \emph{Does robust generalization for standard training also require more data?} If so, there seems to exist some contradictions since other works~\citep{Tsipras2018robustness,su2018robustness} proposed that the robustness may be odds with accuracy. It is well-known that more data can improve the generalization, another natural question needs to be answered: \emph{will robustness be improved or worsen as the data size increases, especially when data is large enough?}
	
	Considering the issues above, we conduct an empirical exploration towards the comprehensive understanding of adversarial examples from three aspects: analyzing the generalization and robustness from limited data to the ``infinite'', task-dependent and model-specific factors, attempting to unify previous research and provide new insights in our explanatory framework. In particular, aiming at answering the aforementioned questions between robustness and generalization with regards to the data size, we investigate the variation of robustness for \emph{standard training} by changing the size of training data, especially achieving the data augmentation based on Auxiliary Classifier GAN (ACGAN)~\citep{odena2016conditional}. It turns out that with the increase of training data, there indeed exists a trade-off relationship that the robustness deteriorates as the generalization performance increases when the training data are limited, however, the robustness starts to improve when the size of data is large enough and finally robust generalization tends to converge to standard generalization, as shown in Figure~\ref{fig:generalization}. 
	This experimental result demonstrates that in limited data regime, adversarially robust generalization for standard training also requires more data. This finding for standard training align with the observation in the adversarial training scenario shown by ~\cite{Tsipras2018robustness,su2018robustness}. However, we further show that the trade-off relationship between generalization and robustness only exists in the restricted training data. When the size of training data is large enough, the trade-off disappears and the classifier can achieve both good  generalization and robustness. To the best of our knowledge, we are the first to reveal the full spectrum of relationship between generalization, robustness and data size for standard training. 
	
	As for the task-dependent factors, we investigate the correlation between the input dimension, number of categories to classify and the robustness, respectively. An interesting finding of our analysis is that the robustness firstly increases and then decreases as the input dimension expands, while it shows an apparent downtrend as the number of categories increases. This discloses the correlation between the complexity of decision boundaries and the vulnerability. For model-specific factors, we validate that the current convolutional neural networks actually have better robustness in comparison with other machine learning methods and expanding network capacity in essence cannot provide real robustness though it can contribute to defense against gradient-based attacks and mitigate transferability. In summary, the contributions of the paper are listed below:
	\begin{itemize}
		\item We provide a systematic analysis on adversarial examples for standard training and unify relevant previous works~\citep{Tsipras2018robustness,su2018robustness,schmidt2018adversarially}, paving a way towards better understanding about adversarial examples.
		\item We present the global relationship between standard generalization and robust generalization for standard training, showing the trade-off relationship in limited data and consistency when data size is large enough.
		\item We validate the influence of task-dependent factors. Increasing the complexity of decision boundaries via increasing input dimensions and number of categories, can make classifier more susceptible to adversarial attacks.
		\item We demonstrate that the current convolutional neural networks have better robustness than traditional ML approaches and reveal that increasing model capacity actually cannot bring real robustness albeit its better robustness against limited attacks and mitigation of transferability. 
	\end{itemize}
	
	\section{A Closer Look at Adversarial Examples}
	The existence of adversarial examples in various machine learning systems demonstrates that the robustness problem is an inherent property of the statistical setup. Here we refine the understanding on adversarial examples. We briefly recap the definition of adversarial examples: crafted indistinguishable examples by adding maliciously constructed perturbations on input data, causing the classifier to produce misclassified predictions. 
	
	
	Define a probability space $(\Omega,\mathcal{F},\mathcal{P})$ and the probability measure $\mathcal{P}$ is called the \textit{population}. The data set $\mathcal{D}$ is viewed as a realization of a random element of this probability space. Due to the imperceptibility in human vision, we assume that both adversarial examples and legitimate examples are sampled from the identical population $\mathcal{P}$ behind albeit the low probability of occurrence for adversarial examples~\cite{nguyen2015deep,yarin2018sufficient}.  
	Due to the randomness of $\mathcal{D}$, the discrepancy between the decision boundary of classifier trained on limited samples and the oracle one based on all population data enables some legitimate data including crafted adversarial examples, misclassified by current imperfect classifier. 
	
	\begin{figure}[t]
		\centering
		\centering\includegraphics[width=.4\textwidth,trim=0 250 0 250,clip]{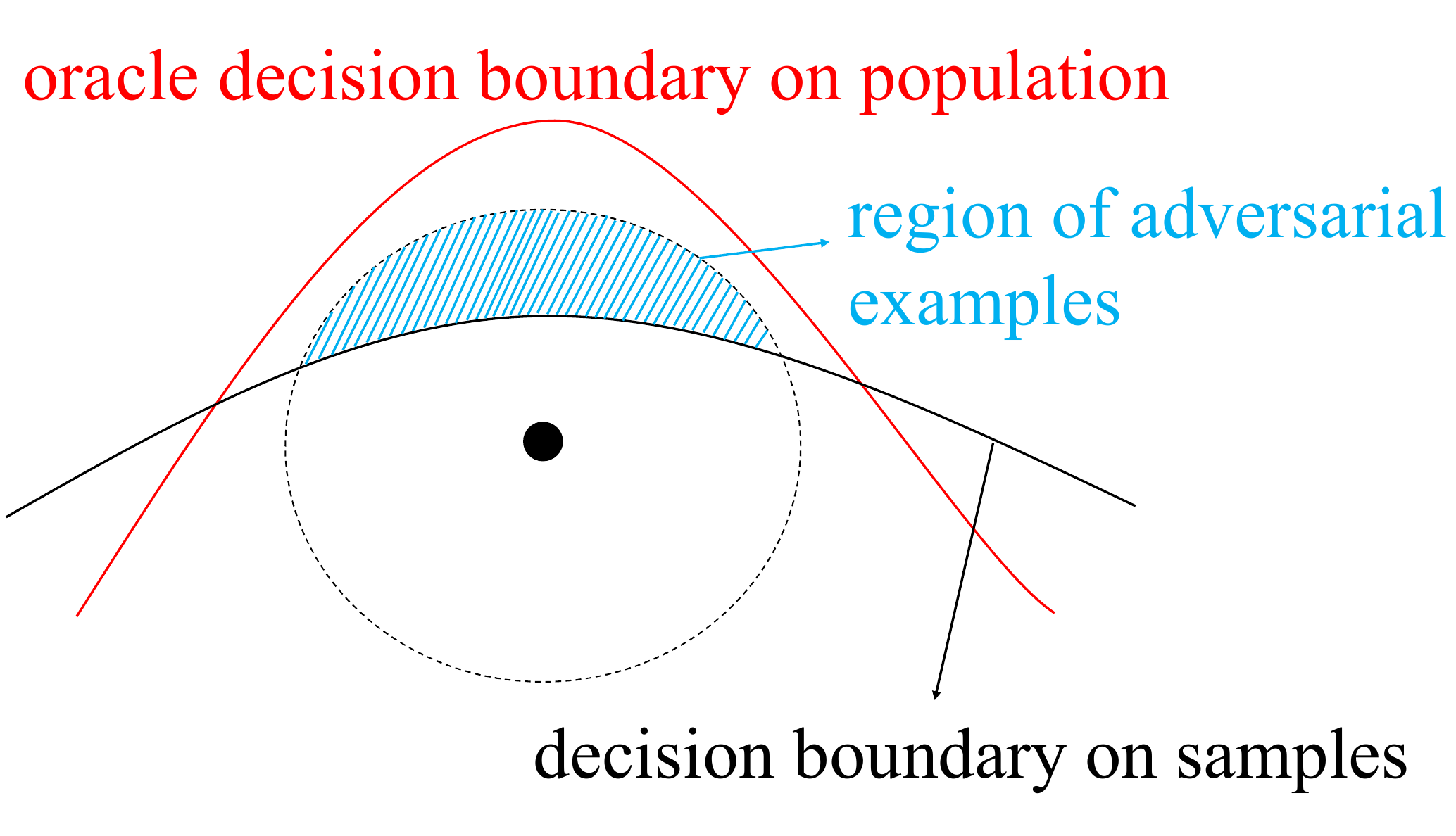}
		\caption{Location of adversarial examples in the proximity of an input example.}
		\label{fig:distribution}
	\end{figure}
	
	More specifically, consider all adversarial examples with restricted perturbations around a correctly-classified example shown in Figure~\ref{fig:distribution}. Due to the discrepancy between the two decision boundaries based on training samples and population, there exists some legitimate examples of a given class in the vicinity of an original example misclassified by existing classifier though they visually should belong to the population data from that class. Based on this comprehension,  we argue that it is the shortage of effective training data that prevents the classifier from including all adjacent examples, particularly those in the blue region in Figure~\ref{fig:distribution}, resulting in the existence of adversarial examples. 
	
	It underlies the core question: \emph{will the robustness be enhanced if we offer sufficient training data to the classifier?} It is our natural expectation that when the size of data is large enough, it would be sufficient to learn robust models. In other words, the generalization and robustness are expected to be consistent with respect to the amount of data. In addition, it is also capable of resolving the inconsistency of relationship between generalization and robustness proposed in the introduction part by validating this hypothesis. 
	
	Besides that, what task and model factors affect the robustness? Is there any effect of the complexity of decision boundaries on the vulnerability of deep neural networks? Are deep neural networks themselves more susceptible to adversarial attacks compared with other machine learning approaches? All of these issues are what we pursue to explore in the following sections. In summary, we investigate these problems through three experimental parts in this work.
	\paragraph{Data size analysis.} Try to uncover the global relationship between standard generalization and robust generalization for standard training with respect to the data size.
	\paragraph{Task-dependent factors analysis.} Attempt to explore the influence of input dimension and number of categories on the robustness of classifier.
	\paragraph{Model-specific factors analysis.} Compare with traditional ML methods to inspect the vulnerability of convolutional neural networks and then investigate the effect of network capacity~\citep{madry2017towards} on the robustness.
	
	\section{Experimental Settings}
	We demonstrate our explanatory framework by performing experiments on several commonly used datasets: MNIST, SVHN, CIFAR10. Fashion-MNIST is an alternative in the first part due to the limited generalization of ACGAN~\citep{odena2016conditional} on SVHN. The experimental setup is as follows.
	\paragraph{Adversarial attacks.} To provide a thorough evaluation of the robustness, various well-known attacks are considered: Fast Gradient Sign Method (FGSM)~\citep{goodfellow6572explaining}, PGD attack~\citep{madry2017towards}, Randomized Fast Gradient Sign Method (RAND+FGSM)~\citep{tramer2017ensemble} and Carlini-Wagner (CW) attack~\citep{carlini2017towards} with $l_2$-norm.
	\paragraph{Model architectures.} For the classifier on MNIST and Fashion MNIST, we adopt the simple architecture with two convolutional layers and three fully-connected layer and for SVHN and CIFAR-10, we consider the standard ResNet18 model. All of our models are trained with identical setting of optimizer for fair comparison and could achieve the state-of-the-art test accuracy on clean data for corresponding datasets.
	\paragraph{Evaluation of robustness.} We consider the original images that are correctly classified to eliminate the influence of standard generalization. Then we evaluate the classification accuracy on the adversarial examples for these correctly classified images and denote it as \textit{Success Defense Rate}.
	
	\section{Data Size Analysis}
	In this section, we will show that in the regime of limited data and data augmentation case, the generalization and robustness of the classifier in the standard training behave differently. In the scenario of limited training data, the generalization and robustness exhibit a trade-off relationship, while in the setting of nearly infinite data, the robustness tends to be consistent with generalization.  
	
	\begin{figure}[t!]
		\centering
		\centering\includegraphics[width=.35\textwidth,trim=50 260 10 260,clip]{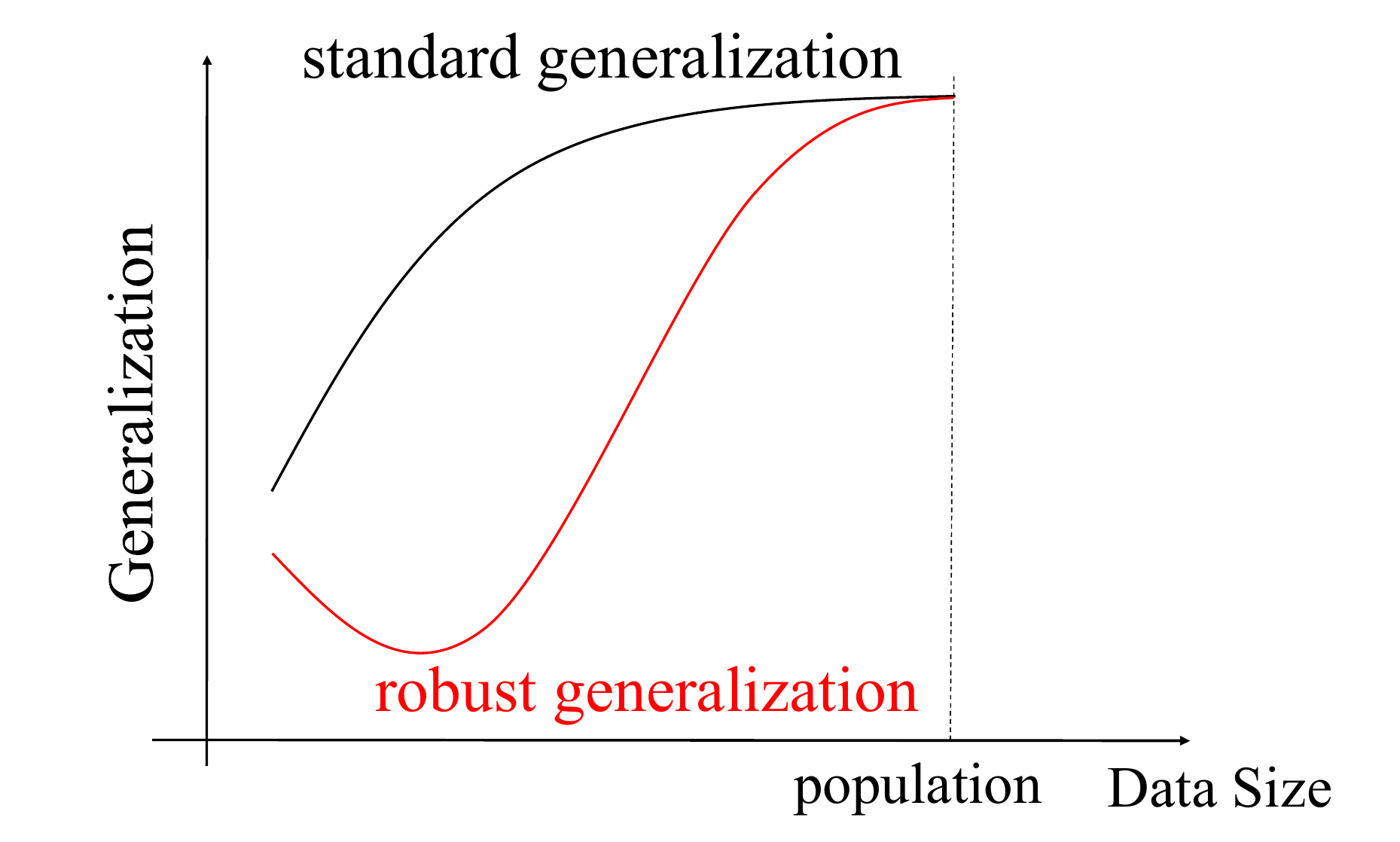}
		\caption{Relationship between standard generalization and robust generalization with respect to the amount of data.}
		\label{fig:generalization}
	\end{figure}
	
	\begin{figure*}[ht]
		\centering
		\centering\includegraphics[width=.52\textwidth,trim=110 0 130 20,clip]{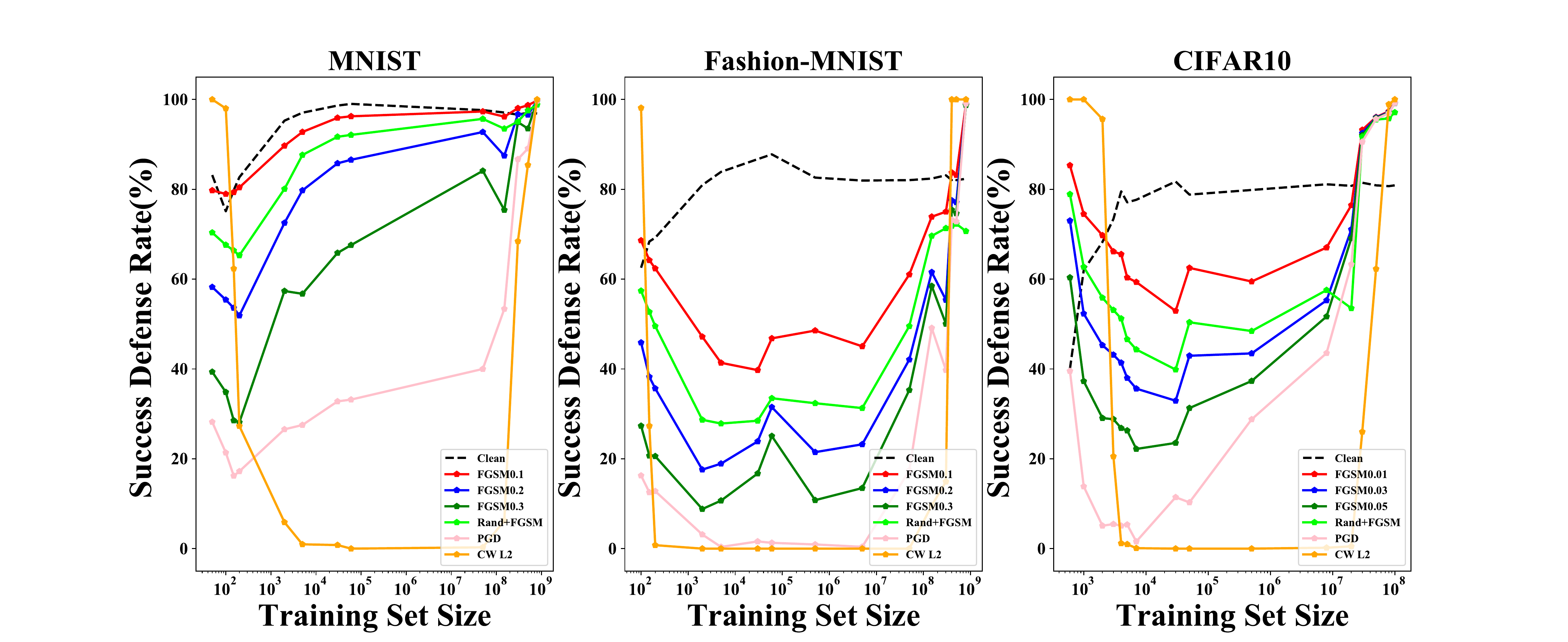}
		\caption{Relationship between generalization and robustness with respect to the data size under different adversarial attacks on MNIST, Fashion-MNIST and CIFAR10. Black dashed line represents the test accuracy and decimals behind FGSM denote the perturbation $\epsilon$.}
		\label{fig:dataaugmentation}
	\end{figure*} 
	
	More specifically, we follow the definition of standard generalization and robust generalization from \citep{schmidt2018adversarially}. In general, standard generalization measures the generalization over the clean test data while robust generalization evaluates the generalization in the adversarial setting where the classification should consider all examples in a perturbation set of original examples. Here we aim to investigate whether adversarially robust generalizaton for standard training also requires more data than standard generalization. If yes, we also need to explore the contradiction behind with the proposals~\citep{su2018robustness,Tsipras2018robustness} that robustness may be at odds with generalization. We demonstrate the hypothesis above and resolve the contradiction by exploring the global  relationship between robust generalization and standard generalization, especially through data augmentation. We sketch this global relationship in Figure~\ref{fig:generalization}.
	
	To verify the relationship between the standard generalization and robust generalization shown in Figure~\ref{fig:generalization}, we investigate the change of generalization and robustness with respect to the size of training data by varying the number of training samples from relatively small to large enough. Generalization is measured by the accuracy on the test set and the robustness is evaluated by \textit{Success Defense Rate}. When the data size is not larger than the existing dataset, we partition its training set with $n$ sub-datasets with a strict inclusion relationship: $A_1 \subset A_2 \subset ... \subset A_n$ and then we train models on each sub-dataset, respectively. When the data size expands further, we utilize the ACGAN~\citep{odena2016conditional}, a version of conditional GAN~\citep{mirza2014conditional}, to model the conditional distribution of data first and then to generate new images to augment the training set. Due to the limitation on precision of ACGAN, we inject clean examples during the training iterations to maintain the accuracy of original classifier. We conduct experiments on MNIST, Fashion-MNIST and CIFAR10 and the  results can be observed in Figure~\ref{fig:dataaugmentation}.
	
	There is a consistent manifestation from Figure~\ref{fig:dataaugmentation} that the robustness evaluated on \textit{Success Defense Rate} decreases first and then increases later as the training data expands across the three datasets. Nevertheless, the turning point of robustness differs when the dataset changes, which is observed earliest for MNIST, then for Fashion-MNIST and last for CIFAR10, just in the ascending order of complexity of datasets. In addition, the result suggests that \textit{Success Defense Rate} under CW attack turns up later than the other attacks, indicating that CW is a much stronger attack so that it requires more data to defense against it, which coordinates with recent experimental conclusions~\citep{carlini2017towards}. As the size of data is  enlarged, \textit{Success Defense Rate} quickly increases to a  high level and then saturates. Although there is a slight decline in the phase of data augmentation due to the limited precision of ACGAN, the apparent uptrend can still be easily observed, providing a strong evidence to our previous statement.
	
	Through the data size analysis via data augmentation based on generative models, we have revealed a global relationship between standard generalization and robust generalization for standard training. The experimental result extends and unifies the viewpoints proposed in \citep{su2018robustness,schmidt2018adversarially,Tsipras2018robustness}:
	
	On one hand, it indeed shows a trade-off relationship between the generalization and robustness for standard training before the turning point occurs. As for the underlying reason, we conjecture that with a limited number of  samples, training on these data helps the classifier to find a better and clearer decision boundary, resulting in high robustness in the initial stage, just as proposed in \citep{anonymous2019how}. As the data size increases, the decision boundary tends to be more complicated and delicate while more training data may lead to higher accuracy, which yields adversarial examples to forge easily. In this case, the generalization and robustness for standard training gradually develop a trade-off relationship.
	
\begin{figure}[htbp]
		\centering
		\centering\includegraphics[width=.48\textwidth,trim=100 10 100 50,clip]{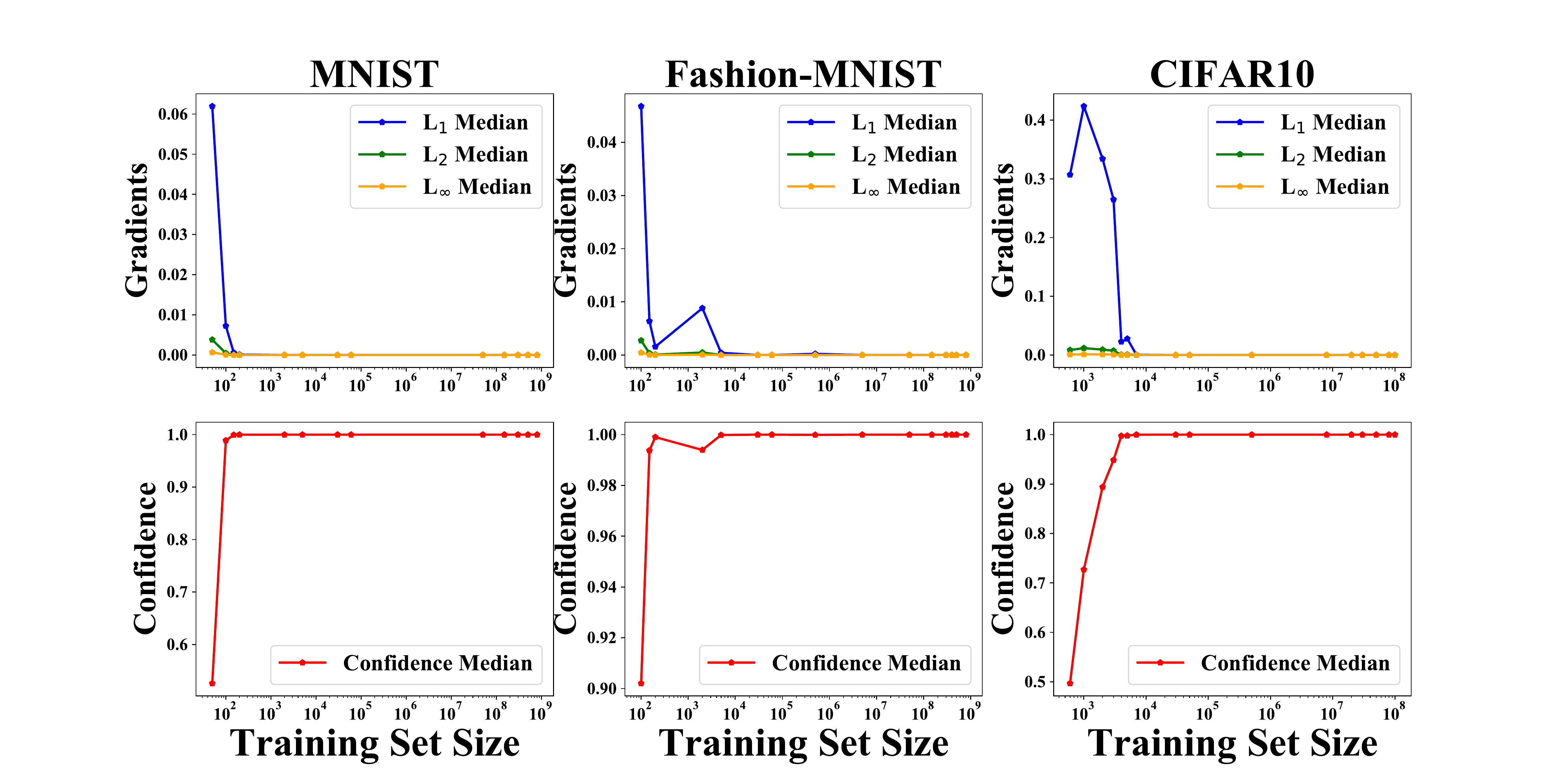}
		\caption{Relationship between median of gradients~(the first row), confidence~(the second row) and the data size on MNIST, Fashion-MNIST and CIFAR10. Median gradients instead of average ones are adopted to avoid the influence of extreme values.}
		\label{fig:grad_confidence}
	\end{figure}

	On the other hand, when the training data is large enough with respect to the complexity of classifier, \textit{Success Defense Rate} ascends to close 100$\%$ where no adversarial attacks can be  successfully crafted, implying that the robustness generalization converges to the standard generalization for standard training. To further explore the reason of the change of robustness, we calculate the magnitude of gradients with respect to the input images under $L_{1}, L_{2}$ and $L_{\infty}$ norms;  and plot the relationship between median of those gradients and the data size, shown in  Figure~\ref{fig:grad_confidence}. An interesting observation is that the magnitude of gradients  gradually vanishes with the increase of the data size. To further probe the cause of gradients vanishing, we exhibit the change of median of confidence measured by maximum softmax probability of network outputs and demonstrate that \emph{it is the saturation of prediction probability  that yields the vanishing of gradients}. More specifically, as the size of data increases, the classifier is capable of including more examples in input space, especially those in the blue region in Figure~\ref{fig:distribution}. In this case, the gap between decision boundary on samples and the oracle one is narrowed, thus improving the robustness as claimed in Session 2. We argue that this type of gradient vanishing is not the phenomenon of \textit{obfuscated gradients} proposed in \citep{athalye2018obfuscated} since our method is not involved in any complex defensive mechanism, on the contrary, it is actually the presentation of real robustness when data size is large enough.
	
	\paragraph{Recommendation on defense} In consideration of the global relationship between robustness and generalization, we speculate there might exist a kind of phenomenon called \emph{robustness trap}, in which the robustness is likely to continue to deteriorate when we augment training data from the existing dataset although the generalization might be improved during this process. Only if the augmented data is large enough can the classifier overcome this trap and its robustness is likely to turn better, which suggests us  establish a more precise generative model to overcome the phenomenon in the future.
	
	\section{Task-Dependent Factors}
	For the classification task-dependent factors, we demonstrate that it is the complexity of decision boundaries that aggravates the problem of adversarial examples. Here we explore the task-dependent factors from two explicit aspects: input dimension and number of categories to classify.
	
	\subsection{Input Dimension}
	For the factor of input dimension, in general we propose that classification tasks with more input dimensions are more vulnerable to adversarial attacks. We verify this proposal in the following.
	
	Previous work~\citep{anonymous2019adversarial} states that in the space of high dimensional data, correctly classified data are very close to misclassified examples, especially for adversarial examples. It adheres to our intuition that the distance between examples is becoming subtler in the high dimensional space, thus aggravating the vulnerability of classifier. \cite{simon2018adversarial}  has proved this influence of input dimension on robustness in the adversarial regularization scenario. In contrast, we conduct our exploration in a more general explanatory framework, getting rid of the gradient-based regularization~\citep{simon2018adversarial} and perform experiments on more datasets with various types of attacks. The observation from our experiments discloses some distinctions with previous work~\citep{simon2018adversarial}, showing that the integral correlation between the input dimension and robustness is not just a simple monotonous one.
	
\begin{figure}[htbp]
		\centering
		\centering\includegraphics[width=.45\textwidth,trim=105 20 105 55,clip]{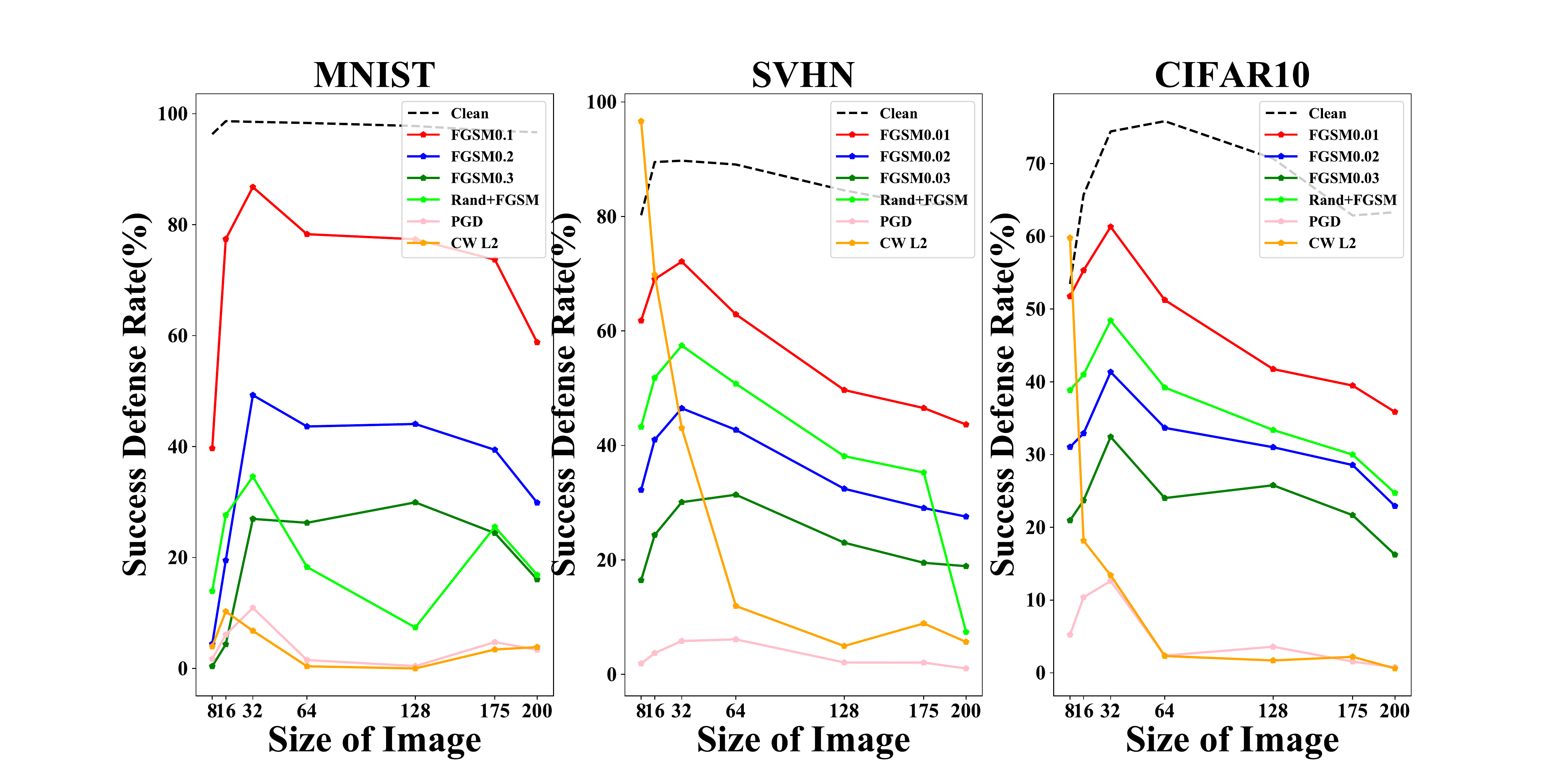}
		\caption{Relationship between the robustness and the input size on MNIST, SVHN and CIFAR10. The input dimension is proportional to the square of input size.}
		\label{fig:dimension}
	\end{figure}

	We implement the experiments by resizing the input images based on bilinear interpolation into different sizes. For the convenience of design and the adaptation of datasets, we apply different networks with similar  architecture on three datasets. For MNIST, we adopt a simple network with one convolutional layer and two fully-connected layers. For SVHN and CIFAR10, we apply networks with three convolutional layers and three fully-connected layers, seven convolutional layers and three fully-connected layers, respectively. The experimental results are shown in Figure~\ref{fig:dimension}. 
	
	In general, the robustness presents a decreasing tendency as the input dimension expands. However, it can be easily observed that except CW attack, the robustness becomes worse when the input dimension is too small. We hypothesize that it is the small input dimensions that makes the classifier under standard training suffer from overfitting, which can be verified by the decline of generalization as well. 
	
	\subsection{Number of Categories to Classify}
	For the factor of number of categories to classify, it is intuitive for us to expect that as the the number of categories grows, the decision boundary will become more complicated, causing the classifier more susceptible to attacks. Thus, we select training data with label from 0 to $n$ and construct several subsets with a strict inclusion relationship by enlarging $n$. Then the networks with different categories to classify are trained on these datasets and evaluated on corresponding robustness under various attacks.
	
	\begin{figure}[hbtp]
		\centering
		\centering\includegraphics[width=.45\textwidth,trim=105 20 105 50,clip]{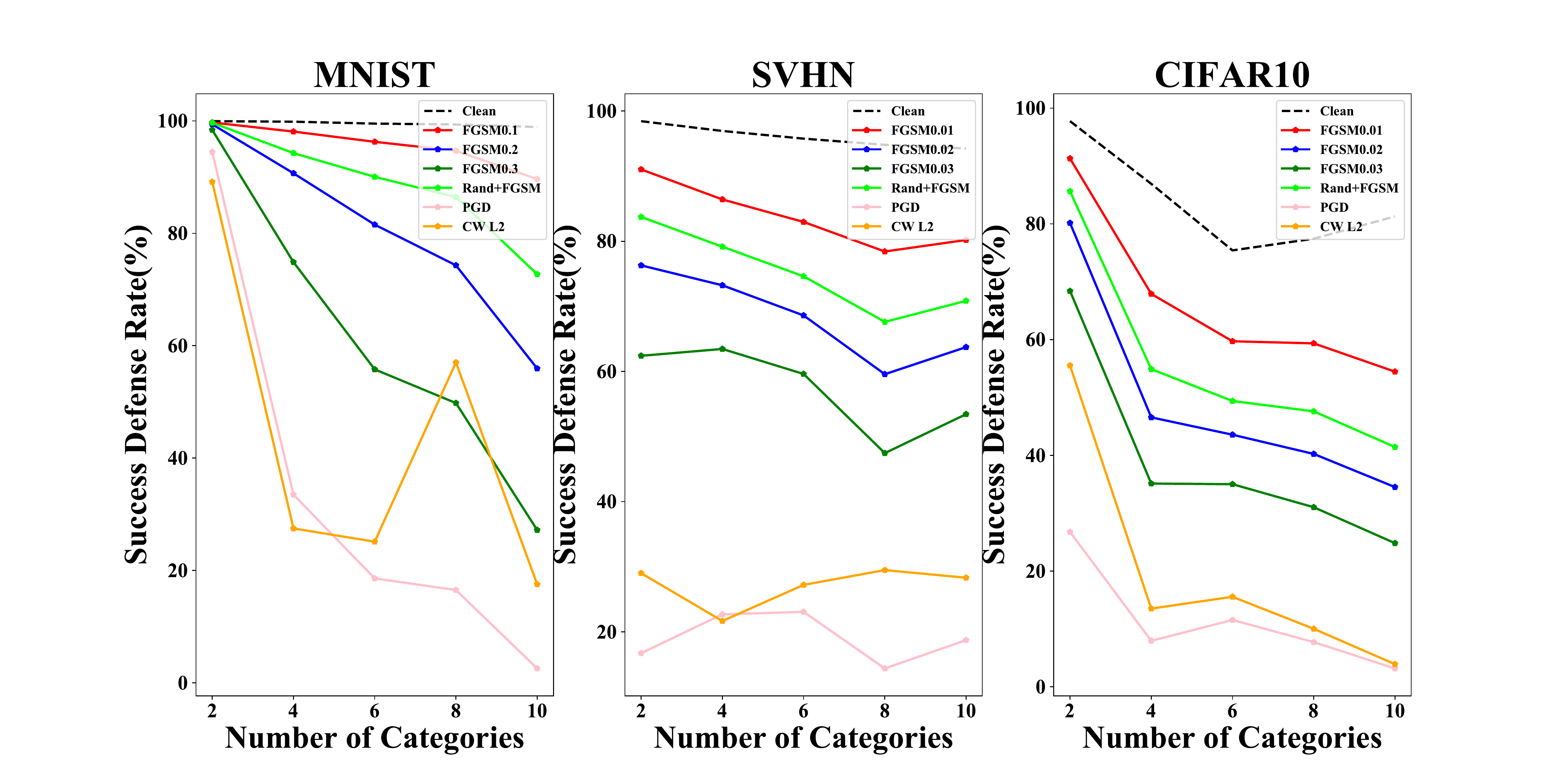}
		\caption{Relationship between robustness and number of categories to classify on MNIST, SVHN and CIFAR10.}
		\label{fig:categories}
	\end{figure}
	
	As illustrated in Figure~\ref{fig:categories}, there is an apparent downtrend of robustness with the increasing of categories to classify though we have to admit that some stochastic effects still exist. For the three datasets, deep neural networks that are required to classify more categories under standard training are more vulnerable to adversarial attacks, which is consistent with our intuition from the perspective of complexity of decision boundaries mentioned before.
	
	\paragraph{Recommendation on defense}
	It is discouraging to admit the fact that deep neural networks in those classification tasks with higher input dimensions and more categories to classify are liable to suffer from adversarial attacks since in practice this is the trend to utilize deep neural networks to tackle  more complicated tasks in the future. Nevertheless, we might consider to combine the models dealing with relative simple tasks to resolve the complex task, probably maintaining a high robustness. In addition, resizing the image to a proper size is beneficial to robustness as well.
	
	\section{Model-Specific Factors}
	For the model-specific factors, on the one hand, we reveal that the robustness of deep neural networks is better than other machine learning models for standard training. On the other hand, we demonstrate that increasing model capacity can help to defend against gradient-based attacks but it actually cannot yield real robustness since they are still fragile faced with alternative optimization-based CW attacks.
	
	\subsection{Comparison with Other Machine Learning Approaches}
	We conduct experiments by comparing the robustness among CNN, LinearSVM and Logistic regression on the three datasets. The CNN has two convolutional layers and two fully-connected layers. Logistic regression adopts standard cross entropy loss and LinearSVM employs a variation of multiclass hinge loss.
	
	\begin{figure}[hbtp]
		\centering
		\centering\includegraphics[width=.45\textwidth,trim=100 0 100 50,clip]{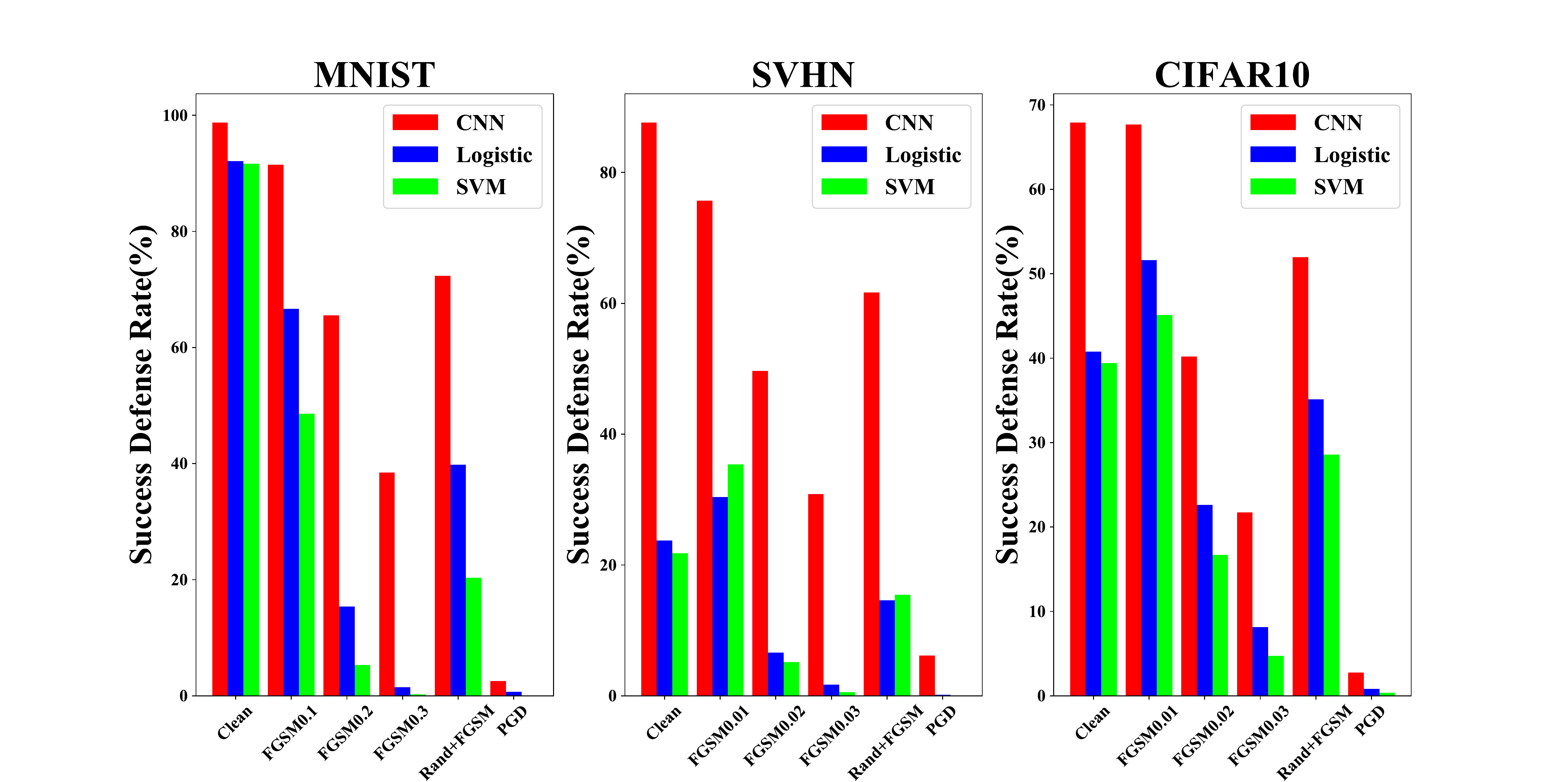}
		\caption{Comparison of robustness among CNN, Logistic regression and LinearSVM on MNIST, SVHN and CIFAR10.}
		\label{fig:model_type}
	\end{figure}
	
	It might be the preconceived notion that deep neural networks are more susceptible to adversarial examples than other traditional machines learning models due to their uncontrollable Lipschitz constants~\citep{zantedeschi2017efficient}. However, our experimental results present evidence that Success Defense Rates of CNN under each attack are consistently superior to the others, suggesting that the robustness of deep neural networks is much better than other typical machines learning systems, as shown in Figure~\ref{fig:model_type}. One thing should be mentioned that although the evaluation of robustness is based on \textit{Success Defense Rate}, which only examine on truly classified examples by given classifier, this index is actually inevitably influenced by generalization in this comparison because LinearSVM and Logistic regression cannot attain the similar generalization performance as CNN especially on SVHN and CIFAR10. We can approximately conclude that CNN has both better generalization and robustness that other machine learning approaches.
	
	\subsection{Model Capacity}
	\cite{madry2017towards} demonstrated that larger model capacity can decrease transferability of adversarial examples, however, we find out increasing model capacity actually cannot bring real robustness by additionally testing CW attack.
	
	\begin{figure}[htbp]
		\centering
		\centering\includegraphics[width=.45\textwidth,trim=100 10 100 50,clip]{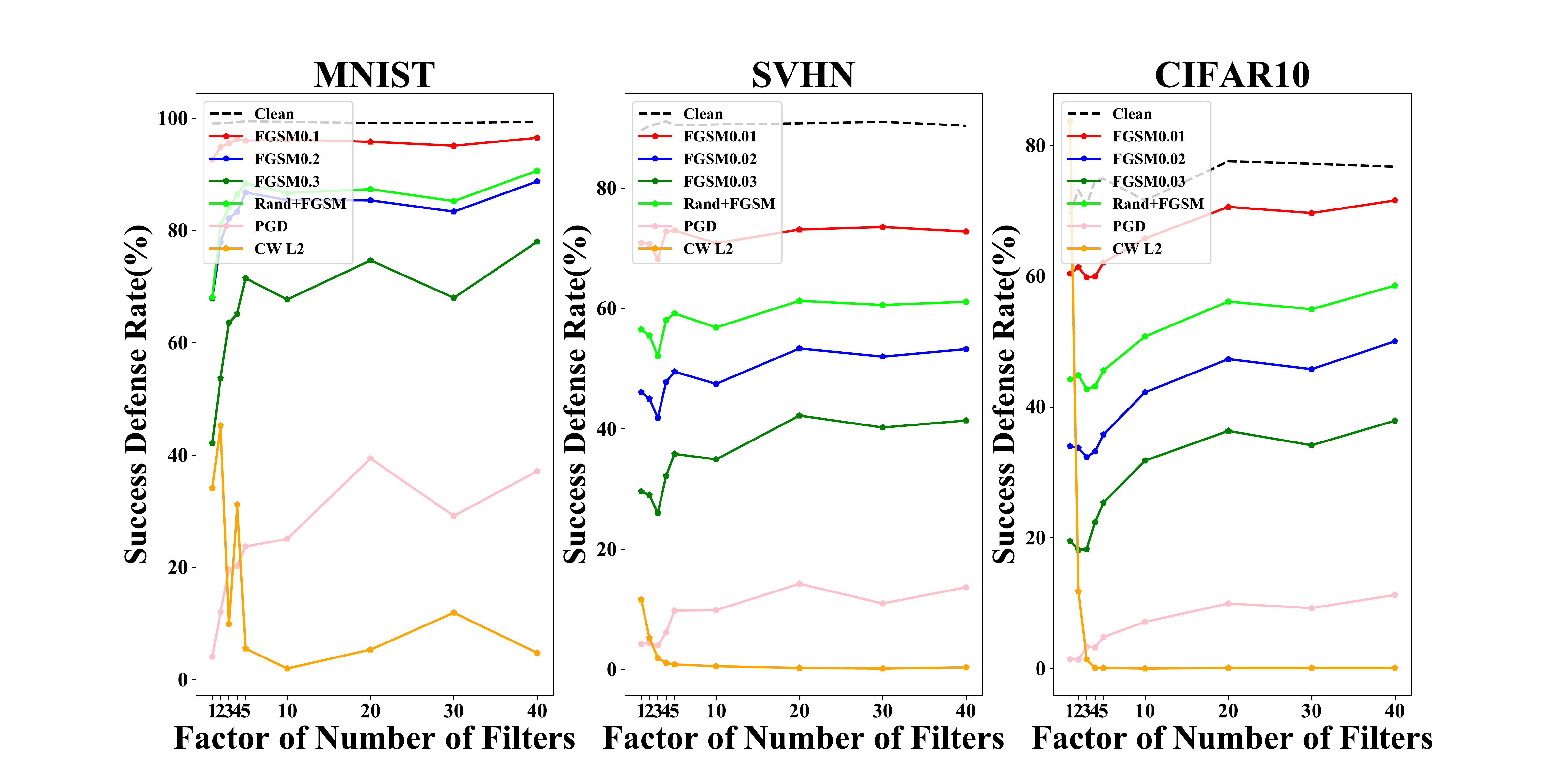}
		\caption{Relationship between the robustness and model capacity. X axis is in log transformation.}
		\label{fig:num_filters}
	\end{figure}
	
	We follow the definition of model capacity in \citep{madry2017towards}, namely the number of filters, and adopt the network architecture with four convolutional layers and three fully-connected layers for the convenience of implementation. To increase the network capacity, we modified the network by incorporating wider layers with different factors $n$, resulting in the enlargement of number of filters with certain magnification. Figure~\ref{fig:num_filters} depicts the trend of robustness with respect to the number of filters. We can observe that Success Defense Rate exhibits an apparent uptrend against gradient-based attacks, i.e. FGSM and PGD attack. \cite{madry2017towards} stated that increasing the network capacity improves the resistance against transfer attacks, which is in accordance with our result since gradient-based attacks are more transferable that CW attack, as proposed by \citep{su2018robustness}. Furthermore, we obtain a more profound conclusion that increasing network capacity actually is unable to improve the real robustness of deep neural networks since the networks are more vulnerable against alternative optimization-based CW attacks~\citep{carlini2017towards}. This also raises an open problem that what are the difference between CW attacks and others.
	
	\begin{figure}[htbp]
		\centering
		\centering\includegraphics[width=.45\textwidth,trim=100 10 100 50,clip]{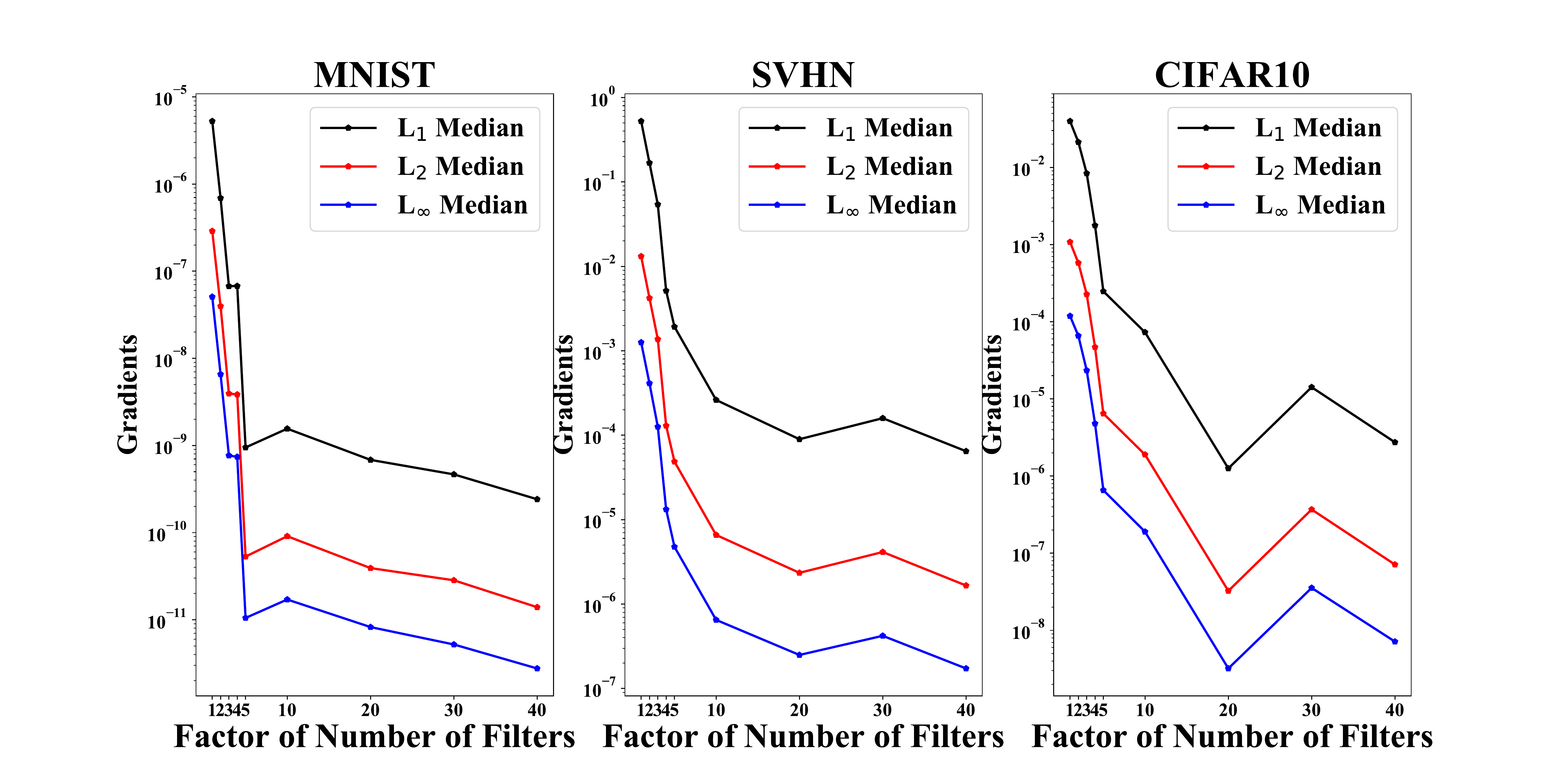}
		\caption{Relationship between model capacity and median of magnitude of gradients with $L_{1}, L_{2}, L_{\infty}$ norms.}
		\label{fig:grad_capacity}
	\end{figure}
	
	Furthermore, we explore the correlation between the magnitude of gradients and model capacity by calculating the median of gradients with different norms. As illustrated in Figure~\ref{fig:grad_capacity}, for all the datasets, it displays an apparent downtrend as the model capacity increases. This implies that the networks with larger capacity have smaller gradients so that they have better robustness against gradient-based attack. However, the reduction of gradients have no improvement on CW attacks that are not directly constructed based on gradients of original loss. More differences between gradient-based attacks and CW attacks are well worth exploring in the future.

	\paragraph{Recommendation on defense}
	Model capacity plays an important role on the robustness but it cannot bring real robustness. Since network architecture is a more crucial factor for robustness, it is a promising direction to design the robust network architectures in the future.
	
	
	\section{Conclusion}
	In this work, we empirically present a systematical study on adversarial examples from three aspects: the amount of data, task-dependent factors and model-specific factors. In particular, we demonstrate that adversarially robust generalization of deep neural networks under standard training also requires more data, and reveal the global relationship between generalization and robustness especially through data augmentation. Then we demonstrate that increasing complexity of decision boundaries will aggravate the vulnerability of deep neural networks from task-dependent factors and demystify the relationship between model-specific factors and robustness. Our analysis sheds light on the systematic understanding of adversarial examples.
	
	\bibliographystyle{named}
	\bibliography{ijcai19}

\begin{thebibliography}{}

\bibitem[\protect\citeauthoryear{Anonymous}{2019a}]{anonymous2019adversarial}
Anonymous.
\newblock Adversarial examples are a natural consequence of test error in
  noise.
\newblock In {\em Submitted to International Conference on Learning
  Representations}, 2019.
\newblock under review.

\bibitem[\protect\citeauthoryear{Anonymous}{2019b}]{anonymous2019how}
Anonymous.
\newblock How training data affect the accuracy and robustness of neural
  networks for image classification.
\newblock In {\em Submitted to International Conference on Learning
  Representations}, 2019.
\newblock under review.

\bibitem[\protect\citeauthoryear{Athalye \bgroup \em et al.\egroup
  }{2018}]{athalye2018obfuscated}
Anish Athalye, Nicholas Carlini, and David Wagner.
\newblock Obfuscated gradients give a false sense of security: Circumventing
  defenses to adversarial examples.
\newblock {\em arXiv preprint arXiv:1802.00420}, 2018.

\bibitem[\protect\citeauthoryear{Buckman \bgroup \em et al.\egroup
  }{2018}]{buckman2018thermometer}
Jacob Buckman, Aurko Roy, Colin Raffel, and Ian Goodfellow.
\newblock Thermometer encoding: One hot way to resist adversarial examples.
\newblock 2018.

\bibitem[\protect\citeauthoryear{Carlini and Wagner}{2017}]{carlini2017towards}
Nicholas Carlini and David Wagner.
\newblock Towards evaluating the robustness of neural networks.
\newblock In {\em 2017 IEEE Symposium on Security and Privacy (SP)}, pages
  39--57. IEEE, 2017.

\bibitem[\protect\citeauthoryear{Dimitris~Tsipras}{2018}]{Tsipras2018robustness}
Logan Engstrom Alexander Turner Aleksander~Madry Dimitris~Tsipras,
  Shibani~Santurkar.
\newblock Robustness may be at odds with accuracy.
\newblock {\em arXiv preprint arXiv:1805.12152}, 2018.

\bibitem[\protect\citeauthoryear{Goodfellow \bgroup \em et al.\egroup
  }{2014}]{goodfellow6572explaining}
Ian~J Goodfellow, Jonathon Shlens, and Christian Szegedy.
\newblock Explaining and harnessing adversarial examples.
\newblock {\em arXiv preprint arXiv:1412.6572}, 2014.

\bibitem[\protect\citeauthoryear{Kurakin \bgroup \em et al.\egroup
  }{2016}]{kurakin2016adversarial}
Alexey Kurakin, Ian Goodfellow, and Samy Bengio.
\newblock Adversarial machine learning at scale.
\newblock {\em arXiv preprint arXiv:1611.01236}, 2016.

\bibitem[\protect\citeauthoryear{Liao \bgroup \em et al.\egroup
  }{2017}]{liao2017defense}
Fangzhou Liao, Ming Liang, Yinpeng Dong, Tianyu Pang, Jun Zhu, and Xiaolin Hu.
\newblock Defense against adversarial attacks using high-level representation
  guided denoiser.
\newblock {\em arXiv preprint arXiv:1712.02976}, 2017.

\bibitem[\protect\citeauthoryear{Madry \bgroup \em et al.\egroup
  }{2017}]{madry2017towards}
Aleksander Madry, Aleksandar Makelov, Ludwig Schmidt, Dimitris Tsipras, and
  Adrian Vladu.
\newblock Towards deep learning models resistant to adversarial attacks.
\newblock {\em arXiv preprint arXiv:1706.06083}, 2017.

\bibitem[\protect\citeauthoryear{Mirza and
  Osindero}{2014}]{mirza2014conditional}
Mehdi Mirza and Simon Osindero.
\newblock Conditional generative adversarial nets.
\newblock {\em arXiv preprint arXiv:1411.1784}, 2014.

\bibitem[\protect\citeauthoryear{Moosavi-Dezfooli \bgroup \em et al.\egroup
  }{2016}]{moosavi2016deepfool}
Seyed-Mohsen Moosavi-Dezfooli, Alhussein Fawzi, and Pascal Frossard.
\newblock Deepfool: a simple and accurate method to fool deep neural networks.
\newblock In {\em Proceedings of the IEEE Conference on Computer Vision and
  Pattern Recognition}, pages 2574--2582, 2016.

\bibitem[\protect\citeauthoryear{Nguyen \bgroup \em et al.\egroup
  }{2015}]{nguyen2015deep}
Anh Nguyen, Jason Yosinski, and Jeff Clune.
\newblock Deep neural networks are easily fooled: High confidence predictions
  for unrecognizable images.
\newblock In {\em Proceedings of the IEEE conference on computer vision and
  pattern recognition}, pages 427--436, 2015.

\bibitem[\protect\citeauthoryear{Odena \bgroup \em et al.\egroup
  }{2016}]{odena2016conditional}
Augustus Odena, Christopher Olah, and Jonathon Shlens.
\newblock Conditional image synthesis with auxiliary classifier gans.
\newblock {\em arXiv preprint arXiv:1610.09585}, 2016.

\bibitem[\protect\citeauthoryear{Papernot \bgroup \em et al.\egroup
  }{2016}]{papernot2016distillation}
Nicolas Papernot, Patrick McDaniel, Xi~Wu, Somesh Jha, and Ananthram Swami.
\newblock Distillation as a defense to adversarial perturbations against deep
  neural networks.
\newblock In {\em 2016 IEEE Symposium on Security and Privacy (SP)}, pages
  582--597. IEEE, 2016.

\bibitem[\protect\citeauthoryear{Samangouei \bgroup \em et al.\egroup
  }{2018}]{samangouei2018defense}
Pouya Samangouei, Maya Kabkab, and Rama Chellappa.
\newblock Defense-gan: Protecting classifiers against adversarial attacks using
  generative models.
\newblock {\em arXiv preprint arXiv:1805.06605}, 2018.

\bibitem[\protect\citeauthoryear{Schmidt \bgroup \em et al.\egroup
  }{2018}]{schmidt2018adversarially}
Ludwig Schmidt, Shibani Santurkar, Dimitris Tsipras, Kunal Talwar, and
  Aleksander Madry.
\newblock Adversarially robust generalization requires more data.
\newblock {\em arXiv preprint arXiv:1804.11285}, 2018.

\bibitem[\protect\citeauthoryear{Simon-Gabriel \bgroup \em et al.\egroup
  }{2018}]{simon2018adversarial}
Carl-Johann Simon-Gabriel, Yann Ollivier, Bernhard Sch{\"o}lkopf, L{\'e}on
  Bottou, and David Lopez-Paz.
\newblock Adversarial vulnerability of neural networks increases with input
  dimension.
\newblock {\em arXiv preprint arXiv:1802.01421}, 2018.

\bibitem[\protect\citeauthoryear{Su \bgroup \em et al.\egroup
  }{2018}]{su2018robustness}
Dong Su, Huan Zhang, Hongge Chen, Jinfeng Yi, Pin-Yu Chen, and Yupeng Gao.
\newblock Is robustness the cost of accuracy?--a comprehensive study on the
  robustness of 18 deep image classification models.
\newblock In {\em Proceedings of the European Conference on Computer Vision
  (ECCV)}, pages 631--648, 2018.

\bibitem[\protect\citeauthoryear{Szegedy \bgroup \em et al.\egroup
  }{2013}]{szegedy2013intriguing}
Christian Szegedy, Wojciech Zaremba, Ilya Sutskever, Joan Bruna, Dumitru Erhan,
  Ian Goodfellow, and Rob Fergus.
\newblock Intriguing properties of neural networks.
\newblock {\em arXiv preprint arXiv:1312.6199}, 2013.

\bibitem[\protect\citeauthoryear{Tanay and Griffin}{2016}]{tanay2016boundary}
Thomas Tanay and Lewis Griffin.
\newblock A boundary tilting persepective on the phenomenon of adversarial
  examples.
\newblock {\em arXiv preprint arXiv:1608.07690}, 2016.

\bibitem[\protect\citeauthoryear{Tram{\`e}r \bgroup \em et al.\egroup
  }{2017}]{tramer2017ensemble}
Florian Tram{\`e}r, Alexey Kurakin, Nicolas Papernot, Ian Goodfellow, Dan
  Boneh, and Patrick McDaniel.
\newblock Ensemble adversarial training: Attacks and defenses.
\newblock {\em arXiv preprint arXiv:1705.07204}, 2017.

\bibitem[\protect\citeauthoryear{Wang \bgroup \em et al.\egroup
  }{2018}]{wang2018one}
Jingkang Wang, Ruoxi Jia, Gerald Friedland, Bo~Li, and Costas Spanos.
\newblock One bit matters: Understanding adversarial examples as the abuse of
  redundancy.
\newblock {\em arXiv preprint arXiv:1810.09650}, 2018.

\bibitem[\protect\citeauthoryear{Yarin~Gal}{2018}]{yarin2018sufficient}
Lewis~Smith Yarin~Gal.
\newblock Sufficient conditions for idealised models to have no adversarial
  examples: a theoretical and empirical study with bayesian neural networks.
\newblock {\em arXiv preprint arXiv:1806.00667}, 2018.

\bibitem[\protect\citeauthoryear{Zantedeschi \bgroup \em et al.\egroup
  }{2017}]{zantedeschi2017efficient}
Valentina Zantedeschi, Maria-Irina Nicolae, and Ambrish Rawat.
\newblock Efficient defenses against adversarial attacks.
\newblock In {\em Proceedings of the 10th ACM Workshop on Artificial
  Intelligence and Security}, pages 39--49. ACM, 2017.

\end{thebibliography}
	
\end{document}